\documentclass[11pt,a4paper]{article}
\usepackage[hyperref]{acl2021}
\usepackage{times}
\usepackage{latexsym}

\usepackage{url}
\usepackage{verbatim} 
\usepackage{graphicx}
\usepackage{enumitem}
\usepackage{microtype}
\usepackage{todonotes}
\usepackage{multirow}
\usepackage{xcolor}
\usepackage{makecell}
\usepackage{soul}
\newcommand\source[1]{\textcolor{blue}{\textsc{\textbf{#1}}}}
\newcommand\target[1]{\textcolor{red}{\textsc{\textbf{#1}}}}

\makeatletter
\renewcommand{\paragraph}{%
  \@startsection{paragraph}{4}%
  {\z@}{.25ex \@plus 1ex \@minus .2ex}{-1em}%
  {\normalfont\normalsize\bfseries}%
}
\makeatother

\aclfinalcopy
\def\aclpaperid{2356} 

\title{Metaphor Generation with Conceptual Mappings}

\author{Kevin Stowe$^1$, Tuhin Chakrabarty$^2$, Nanyun Peng$^3$ \\
\textbf{Smaranda Muresan$^2$, Iryna Gurevych$^1$} \\ 
$^1$Ubiquitous Knowledge Processing Lab, Technical University of Darmstadt \\ 
\url{https://www.informatik.tu-darmstadt.de/ukp/} \\
$^2$Columbia University, \url{{tuhin.chakr,smara}@cs.columbia.edu} \\
$^3$University of California Los Angeles, \url{violetpeng@cs.ucla.edu}
}
\date{}

\begin{document}
\maketitle
\begin{abstract}

Generating metaphors is a difficult task as it requires understanding nuanced relationships between abstract concepts. In this paper, we aim to generate a metaphoric sentence given a literal expression by replacing relevant verbs. Guided by conceptual metaphor theory, we propose to control the generation process by encoding conceptual mappings between cognitive domains to generate meaningful metaphoric expressions. To achieve this, we develop two methods: 1) using FrameNet-based embeddings to learn mappings between domains and applying them at the lexical level (CM-Lex), and 2) deriving source/target pairs to train a controlled seq-to-seq generation model (CM-BART). 
We assess our methods through automatic and human evaluation for basic metaphoricity and conceptual metaphor presence. We show that the unsupervised CM-Lex model is competitive with recent deep learning metaphor generation systems, and CM-BART outperforms all other models both in automatic and human evaluations.\footnote{All code, models, and data are made available at: \url{https://github.com/UKPLab/acl2021-metaphor-generation-conceptual}}
\end{abstract}

\section{Introduction}

Recent neural models have led to important progress
in natural language generation (NLG) tasks. While pre-trained models have facilitated advances in many areas of generation, the field of metaphor generation remains relatively unexplored. Moreover, the few existing deep learning models for metaphor generation \cite{yu2019,stowe2020, chakrabarty2020} lack any conceptualization of the meaning of the metaphors. 

        \begin{figure}[t]
    \includegraphics[width=.45\textwidth]{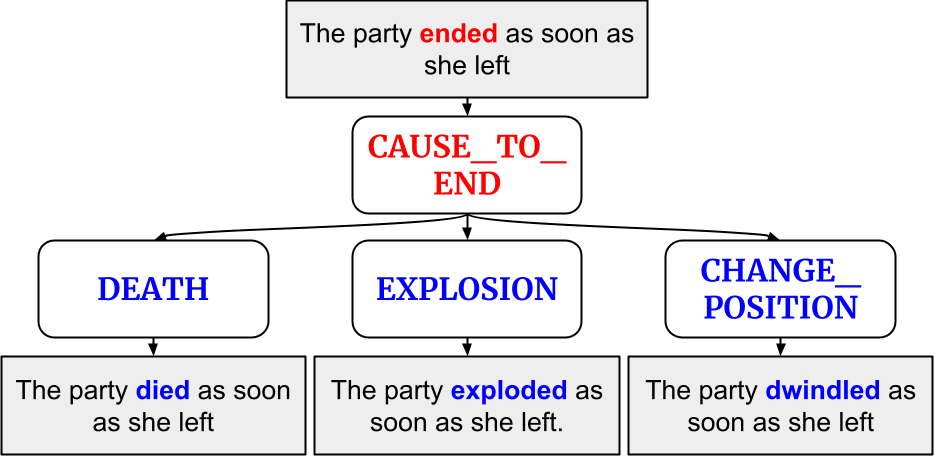}
    \centering
    \vspace{-.5em}
    \caption{\label{fig:main} Metaphor generation guided by conceptual metaphors. Given a literal input, we can generate metaphoric outputs based on different mappings between conceptual domains.}
    \vspace{-1.5em}
    \end{figure}

This work proposes the first step towards metaphor generation informed by the conceptual metaphor theory (CMT) \cite{lakoff1980,lakoff1993,reddy1993}. CMT holds that we use conceptual mappings between domains (conceptual structures that group related concepts) to generate linguistic metaphors.\footnote{``Domains'' are also often referred to as ``image schema'', ``frames'', ``scenes'', and more; see \newcite{kovecses2020}} Metaphoric mappings consist of a \textit{source} and a \textit{target} conceptual domain. 
The \textit{source} domain is the conceptual domain from which we draw the \textit{metaphorical} expressions, while the \textit{target} domain is the conceptual domain that we try to understand. 
A classical mapping is \target{argument} \textsc{is} \source{war}, in which we conceptualize the \textit{target} argumentation domain as the more concrete \textit{source} domain of war:
    
    \begin{itemize}[noitemsep]
        \item They \textbf{fought} against the contract. 
        \item They \textbf{defended} their new proposal.
         \vspace{-1ex}
    \end{itemize}

We focus on verbs, as they are often the key component of metaphoric expressions \cite{steen2010,martin2006}. When used metaphorically, verbs typically evoke source domains (e.g. \textbf{fought}, \textbf{defended} in the above examples): they are concrete, and are used to understand more abstract targets (i.e., argumentation verbs such as \textit{argued}, \textit{supported}) via conceptual mappings \cite{sullivan2013}.

We propose a novel framework for metaphor generation informed by conceptual metaphor theory. 
Given a \textit{literal} input sentence that evokes a \textit{target domain} we generate \textit{metaphoric} sentences that evoke desired corresponding \textit{source} domain(s).\footnote{We note that this source and target terminology used here is opposite to that in machine translation.} For example, given the literal sentence \textit{The party ended as soon as she left} evoking the target domain \target{cause\_to\_end}, we can apply a variety of conceptual mappings to generate different metaphoric outputs evoking different source domains (see Figure \ref{fig:main}). 
This allows us to generate metaphoric expressions that match known metaphoric mappings, as well as generating from unseen mappings to explore novel metaphors. Our contributions are:

\begin{itemize} [noitemsep]
\item Two metaphor generation models grounded in CMT: 1) An unsupervised lexical model relying on frame embeddings learned from Framenet (CM-Lex, Section \ref{sec:framemethod}) and 2) a BART \cite{lewis2020} model encoding source/target domain information through fine-tuning  (CM-BART, Section \ref{sec:bartmethod}).
\item Two metaphor generation tasks: 1) generate metaphoric expressions from known concept mappings, for which we provide gold standard test data, and 2) generate novel expressions from unknown metaphors using rare and unseen mappings (Section \ref{sec:experiments}). 


\item A thorough evaluation using both automatic and human evaluations (Section \ref{sec:results}). We show that our CM-BART model improves over all others in terms of metaphoricity (by $\geq 7\%$) and domain evocation (by $\geq 33\%$), and CM-Lex is competitive with previous neural models on metaphoricity while outperforming them on domain evocation (by $\geq 13\%$). 
\end{itemize}

\section{Task Definition}
Traditional metaphor generation models focus only on whether the generated output is in some way ``metaphoric'' or not. This ignores the semantic and cognitive properties inherent in metaphoricity. These models can, to some degree, generate metaphors given a literal input, but these outputs often do not evoke the intended metaphor. 
    

    
    Controlled metaphor generation yields critical benefits over these uncontrolled systems. 
    For sentences in context, having metaphors that are consistent with the text is essential for natural understanding. Also, metaphors are not only used to express human knowledge, but can also help shape our understanding of the world: having fine-grained control over the generation process allows us to explore novel metaphoric mappings and perhaps improve our understanding of the related domains.
    
    To achieve controlled metaphor generation, we define our task as follows: given \textit{a literal input} sentence which evokes \textit{a target domain} and \textit{an intended conceptual mapping}, generate a metaphoric sentence such that it evokes a desired source domain. 
    Thus, our generation models receive three inputs: 1) a literal input sentence (\textit{They \textbf{argued} against the contract}), 2) the target domain evoked by the literal input (\target{argument}) and 3) the desired source domain (\source{war}) for the metaphorical sentence. The output is a metaphorical sentence which evokes the intended mapping (\textit{They \textbf{fought} against the contract})
    

\section{Methods}
    We experiment with two general categories for generation. First, following previous work in metaphor generation and interpretation \cite{mao2018,stowe2020}, we implement lexical methods for replacement, identifying relevant verbs and replacing them with potential candidates for evoking particular mappings. Second, we experiment with deep learning models, employing controlled sequence-to-sequence generation.

     \subsection{CM-Lex}
     \label{sec:framemethod}
     
     Metaphor generation can be conceptualized as finding key words and replacing them with metaphoric counterparts. This can be done by employing vector spaces, identifying the word most likely to fit in an appropriate context and subjecting them to some constraints of metaphoricity. We build on this paradigm by incorporating facets of conceptual metaphor theory.
     
     Our procedure is as follows: we learn a joint embedded representations for domains and lexical items. We then use the linear transformation between two domains as a mapping, which can be applied to input words from the target domain to generate a word from the source domain.
     As a proxy for domains,  we utilize FrameNet \cite{baker1998}, which contains semantic frames along with the set of lexical units that evoke them. Frames can be defined as related systems of concepts \cite{fillmore1982}, which is exchangeable with the term ``domain'' used in conceptual metaphor theory \cite{croft2004}. Thus, we consider the transformation from one frame to another as a proxy for a conceptual metaphoric mapping.
     
     We first train FrameNet frame embeddings and employ evaluation metrics to ensure their quality. We then apply transformations between domains to literal verbs to generate metaphors grounded in conceptual metaphor theory.
     
     \subsubsection{Learning Frame Embeddings}
     In order to exploit FrameNet frames as conceptual domains, we will embed them in vector space.  While lexical and contextualized embeddings have proven effective, the field of embedding concepts from lexical resources is less well explored \cite{sikos2018,alhoshan2019}. These methods involve tagging raw corpora using automatic FrameNet parsing and then inputting some combination of the original text and the FrameNet information into standard embedding algorithms.
     
    To train and evaluate frame embeddings, we use 211k sentences of Gold annotations used to train the \texttt{Open-SESAME} parser \cite{swayamdipta2017}, along with a variety of other automatically tagged datasets: 250k individual sentence from the Gutenberg Poetry Corpus \cite{jacobs2018}, 17k from various fiction section of the Brown Corpus \cite{francis1979}, and 80k sentences randomly selected from Wikipedia. 
    From this, we extract a 5-word context window for each verb, creating 1.8M verb instances. We then replace the focus verb with its FrameNet frame label (either provided in the Gold data, or tagged via the parser), and train embedding models on the resulting data. This yields joint embedding spaces that contain both common words and FrameNet frame embeddings.
  
    We define two intrinsic metrics to evaluate the quality of our produced embeddings to enable fine-tuning and validation.
    First,  following \newcite{sikos2018}, we can evaluate quality based on the words that evoke that Frame. FrameNet gives a set of lexical units (LUs) that evoke each frame $f$. We calculate the lexical similarity by taking the distance from the mean embedding of ``local'' words ($w \in f$) to the mean embedding of a random sample $k$ of ``distant'' words ($w \not\in f$):
    
    $lex(f) = \sum\limits_{w \in f} \frac{cos(E_w, E_f)}{|f|} - \sum\limits_{w \not\in f}^k \frac{cos(E_w, E_f)}{k}$

    This lexical metric ($lex$) is evaluates whether the frame embedding is similar to words within its frame and dissimilar to those without. 
    
    FrameNet also contains linking relations between frames (eg. \texttt{used-by}, \texttt{uses}), yielding a hierarchy of connected frames. Starting with the assumption that frames connected in the structure should be more similar, we also calculate a structural similarity metric $str$. We follow the same process as above, taking the distance between the mean embedding of the local frames $n \in N$, where $N$ is the immediate neighbors of $f$, to the mean embedding of a sample $k$ of distant frames $n \notin N$.
    
    $str(f) = \sum\limits_{n \in N} \frac{cos(E_n, E_f)}{|N|} - \sum\limits_{n \not\in N}^k \frac{cos(E_n, E_f)}{k}$
    
    We experiment with three lexical embeddings models: word2vec skip-gram \cite{mikolov2013}, Glove \cite{pennington2014}, and FastText \cite{bojanowski2017}. We experiment with 50, 100, and 300 dimensional representations; we find the 50 dimensional word2vec embeddings perform best for both evaluation metrics.\footnote{For full frame embedding evaluation, see Appendix \ref{app:metrics}.}

    \subsubsection{Embedding Mappings}
    \label{sec:mappings}
    To apply these embeddings to generate metaphors based on conceptual mappings, we learn mappings between frames and apply the mappings directly to lexical items to facilitate lexical replacement.
    
\begin{figure}[t]
        \small
    \includegraphics[width=.45\textwidth]{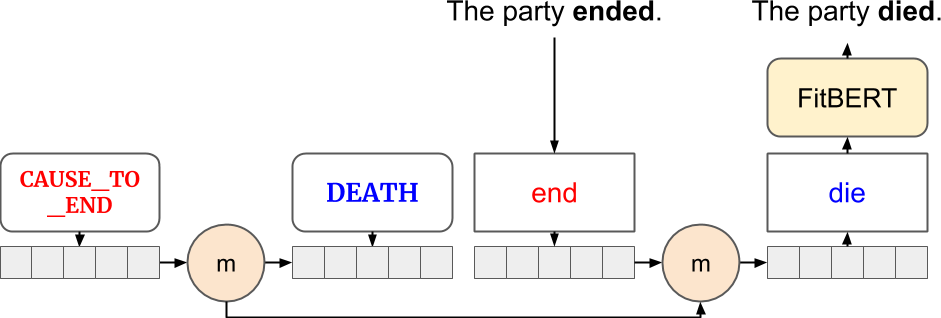}
    \centering
    \vspace{-.5em}
    \caption{\label{fig:lex} Lexical generation process}
    \vspace{-1.5em}
    \end{figure}

    We define a mapping $m$ as the pointwise distance between the target frame embedding and the source frame embedding. Following the approach for learning connections between concrete and poetic themes of \newcite{gagliano2016}, we sum the embedding of the target verb and the mapping $m$ for the selected conceptual mapping, and select the most similar word to the resulting vector. This word is then delemmatized using \texttt{fitbert} \cite{haven2019} and inserted into the original sentence (Figure \ref{fig:lex}). 
    
    Note that these resulting words are generated without context, as they rely only on the input word and the conceptual mappings. This approach has benefits: we require no labeled metaphor data, using only embeddings trained on FrameNet-tagged corpora. However, ignoring context is likely detrimental. In order to better use contextual information, we explore state-of-the-art sequence-to-sequence modeling.

    \begin{table*}[ht]
        \centering
        \small
        \begin{tabular}{c|c|c|c}
       Literal (filled from LM) & Target Frame & Metaphoric (original) & Source Frame\\
        \hline 
        \hline
            That tyranny is \textbf{destroyed} & \target{destruction} & That tyranny is \textbf{slain} & \source{killing} \\
        The house where love had \textbf{ended} & \target{cause\_to\_end} & The house where love had \textbf{died} & \source{death} \\
        As the moments \textbf{passed} on & \target{process\_end} & As the moments \textbf{roll} on & \source{cause\_motion} \\
        What I \textbf{learned} my senses fraught & \target{coming\_to\_believe} & What I \textbf{bear} my senses fraught & \source{bringing} \\
        \end{tabular}
        \caption{Sample of extracted pairs from the data collection process.}
        \label{tab:data}
    \vspace{-3ex}
    \end{table*}

    \subsection{CM-BART}
    \label{sec:bartmethod}
    For sequence-to-sequence learning, we fine-tune a pre-trained BART model \cite{lewis2020}, adding source and target information to guide generation towards the intended metaphors. We first outline a procedure for generating semi-supervised paired data, then detail the training and generation process.

    \subsubsection{Method for Creating  Parallel Data} 
    \label{sec:data-creation}

    In order to train sequence-to-sequence models for metaphor generation, we require large scale parallel corpora. We follow the approach of \citet{chakrabarty2021} and build a corpus of literal/metaphoric paraphrases by starting with the Gutenberg Poetry corpus \cite{jacobs2018}, identifying and masking metaphoric verbs, and replacing them with infilling from a language model. We use  a BERT-based metaphor classification model trained on the VUA metaphor corpus \cite{steen2010} to identify metaphoric verbs in a sentence (i.e ``died'' in \textit{The house where love had \textbf{died}}). Then we convert it to a literal sentence \textit{(The house where love had ended)} using infillings from pre-trained BERT \cite{devlin2019}. 
    
    To ensure the literal sentence with replacements convey the same semantic meaning as the metaphorical sentence they are then filtered using symbolic meaning (\textit{SymbolOf} relation) obtained from COMET \cite{comet}, a GPT based language model fine-tuned on ConceptNet \cite{conceptnet}. COMET returns top 5 symbolic beams of \textit{(loss, loneliness, despair, sadness and sorrow)} for the sentence ``The house where love had \textbf{died}" whereas it replaces \textit{sorrow} with \textit{life} for the literal version. While \citet{chakrabarty2021} filter down to only those candidates with an exact match between the top 5 symbolic beams for the literal and metaphorical sentences returned by the COMET model, we ease the restriction to cases where at least four of five symbols are the same. 
    
    In order to learn more direct metaphoric information from this data, we additionally tag each sentence with FrameNet frames using the \texttt{Open-SESAME} parser \cite{swayamdipta2017}. We extract each pair in which both the focus word in the literal, target-domain sentence and the metaphoric, source-domain sentence are assigned a FrameNet frame. We then make the assumption that the relation between the frames for the source and target domains reflects a metaphoric mapping. This then yields a dataset of paired sentences for which we have a metaphoric mapping between domains based on FrameNet for the focus verbs.
    
     Samples of the created data are shown in Table \ref{tab:data}. In total this process yields 248k sentences spanning 8.5k unique mappings between FrameNet frames. Each pair comprises a literal and metaphoric sentence, along with the literal target frame and the metaphoric source frame. From these we can directly train a sequence to sequence model for conceptual metaphor-based generation. 

    \subsubsection{Models}
    We fine-tune BART \cite{lewis2020}, a pre-trained conditional language model that combines bidirectional and auto-regressive transformers, on the created parallel corpora described in Section \ref{sec:data-creation}. We incorporate representations of the frame information to allow this model to control for the metaphoric mappings evoked. 
        
    To transform a literal sentence from a given target domain to a metaphorical sentence evoking a specific source domain, we incorporate both target and source domains (as FrameNet frames) into the textual representation as a control code, following the work of \newcite{schiller2020aspect} who used this procedure for Argument Generation. Following the example from Figure \ref{fig:main}, the input literal text fed to the BART encoder would be:
        
        \begin{itemize}
            \item \source{death}  $\langle EOT \rangle$ The party  $\langle V \rangle$ ended : \target{cause\_to\_end} $\langle V \rangle$ as soon as she left.
        \end{itemize}
    where $\langle EOT \rangle$ and $\langle V \rangle$ are delimiters, \source{death} is the source frame, and \target{cause\_to\_end} the target frame. The decoding target is the metaphoric text ``The party \textbf{died} as soon as she left'',  which evokes the \target{cause\_to\_end} \textsc{is} \source{death} mapping.
    
    Note that our training data differs only at the level of a single verb. We use the generative BART seq2seq model to generate metaphoric paraphrases, but due to the nature of the training data and the importance of verbs in metaphoric expressions, this is often realized in the output as lexical replacement.
        
    Post fine-tuning, we use top-k (k=5) sampling \cite{fan2018} to generate metaphors conditioned on the input literal sentence and source and target domains for the required metaphoric mapping.\footnote{Full parameter tuning outlined in Appendix \ref{app:params}.} We evaluate the lexical model (\textbf{CM-Lex}) and the sequence-to-sequence model (\textbf{CM-BART}) under two experimental settings.

\section{Experimental Setup}
    \label{sec:experiments}
    We evaluate our metaphor generation methods against two previous approaches to metaphoric paraphrase generation: the \textbf{\textsc{mermaid}} system \cite{chakrabarty2021} and the metaphor masking model (\textbf{MetMask}) \cite{stowe2020}. We explore two tasks: generating against gold standard metaphoric expressions, and using rare and unseen metaphoric mappings. For the former, we build a gold test set of metaphoric paraphrases that evoke a particular source/target mapping. For the latter, we apply a variety of source/target mappings to literal inputs for which we do not have gold outputs.

    \subsection{Building a Test Set}
    \label{sec:testset}
    For a test set, we use the same procedure as our data collection approach from Section \ref{sec:data-creation}. We apply this procedure to two datasets: a sample of the Gutenberg Poetry Corpus and a sample of fiction from the Brown Corpus \cite{francis1979}. This generates an initial set of literal/metaphoric pairs. We also tagged the pairs from \newcite{mohammad2016} with FrameNet tags, as these generally contain novel, well-formed metaphors. These three datasets each have different properties with regard to metaphor. The \textbf{Gutenberg Poetry corpus} has consistent, novel metaphors, but often unconventional syntactic constructions, due to the poetic nature of the text. The \textbf{Mohammad 2016} corpus contains manually constructed metaphors which are novel, following relatively basic syntactic patterns.  The \textbf{Brown Corpus} is standard fiction texts, so the metaphors within tend to be very conventional.
    
    
    From these sources, we draw pairs randomly, checking that they reflect strong literal/metaphoric paraphrases until we obtain 50 instances from each set. Each pair is tagged with FrameNet frames for the focus verbs, which comprise the metaphoric mapping.\footnote{In 22 cases, parsing errors in FrameNet frames were manually corrected.} For the Brown corpus, metaphoric expressions were relatively rare, and thus valid pairings were sparse: to overcome this, we manually modified 11 of the expressions to evoke the appropriate metaphoric mappings. 
    
    In total this process yielded 150 literal/metaphoric pairs, along with the source and target frames that they evoke. 
    We use this dataset to evaluate generating metaphors based on mappings with gold standard outputs, using both automatic and human-based evaluations.  
    
\subsection{Expanding to Unknown Metaphors}
To explore the flexibility of the system developed in this study, we also evaluate them for generation of metaphoric expressions that are not directly linked to gold literal/metaphoric pairs. For this, we begin with our 150 pairs from above, but consider only the literal sentence and the evoked target domain. 
For each sentence, we generate two source domains that could potentially map to the target. These are selected in order to identify rare and unseen mappings based on the observed mappings in our training data. For \textbf{rare} mappings we select a source domain at random from the mappings with the \textit{median} frequency for a given target domain. For \textbf{unseen} mappings we select a source domain at random from the FrameNet frames that are \textit{never} used as a source for the given target domain. 

This set contains only the tuple (input sentence, target domain, source domain) needed as input to our models; we do not have gold generated metaphorical utterances. Thus, on this set we will only perform human-based evaluation of the quality of the generated metaphors.  

\subsection{Automatic Evaluation Metrics}
   Word overlap metrics (eg. BLEU, ROUGE) are inherently weak for this task, as these sentences inherently have high overlaps. So instead, we employ semantic distance metrics. We generate sentence embeddings using SBERT\footnote{Specifically using the \texttt{roberta-large} model, which shows the best performance for sentence similarity tasks.} \cite{reimers2019} for each of our components: the literal input $L$, the original gold metaphoric expression $M$, and the generated output $G$.
    
    \subsubsection{Distance from Gold Metaphor ($dis$)}
        The generated metaphoric expressions should match the semantics of the original gold metaphor.
        We can evaluate this using the cosine distance, here between $M$ and $G$. As SBERT embeddings have been shown to reflect semantic similarity and entailment between paired sentences, this metric should be capable of capturing whether the generated metaphoric expression matches the gold.
        
    \subsubsection{Relational distance ($rel$)}
        Assuming that conceptual metaphoric mappings are responsible for the connecting of meaning between our literal and metaphoric sentences, we would also expect there to be a relation that holds between the original literal input $L$ and metaphoric output $M$. This relation should also hold between the $L$ and the generated metaphor $G$. As a simple metric we can employ cosine distance: we aim for minimizing the distance between $cos(L, M)$ between $cos(L, G)$.

    Finally, we include the percentage of times the model produced the exact gold output. 
    
    \section{Results and Analysis} 
    \label{sec:results}
    Results for automatic evaluation on the 150 gold metaphors are shown in Table \ref{tab:auto-eval}. Note that we cannot automatically evaluate against rare or unseen metaphoric mappings, as we lack gold metaphors.

          \begin{table}[t]
          \small
        \centering
        \begin{tabular}{|c|c|c|c|c|}
        \hline
        \textbf{Model} & $dis$ & $rel$ & $mean$  & \%= \\
        \hline
        MetMask & .191 & .094 & .143 & .087 \\
        \textsc{mermaid}  & .147 & .087 & .117 & .133 \\
        \hline
        CM-Lex  & .151 & .086 & .122 & .107 \\
       
        CM-BART  & \textbf{.085} & \textbf{.047} & \textbf{.066} & \textbf{.293}\\

        \hline
        \end{tabular}
        \caption{Automatic evaluation for metaphor generation systems. \textbf{\%=} indicates the percentage that matched the gold metaphor exactly.}
        \label{tab:auto-eval}
        \vspace{-1.5em}
    \end{table}

    The CM-Lex model is competitive with the best neural baseline, which is encouraging. This shows that simply incorporating basic understanding of conceptual mappings can be a powerful tool for metaphor generation. The CM-BART yields the best automatic performance over all metrics, significantly outperforming all other models ($p < .01$, paired t-test.).  

       Automatic metrics allow us to quickly prototype metaphoric generation systems based in conceptual metaphor theory. However, they rely on SBERT and inherit the biases and weaknesses therein. We also perform human evaluations, against both the gold test data and the set of rare and unseen mappings.

    \subsection{Human Evaluation}
        For human evaluation, we defined two objectives. First, we aim to capture the \textbf{metaphoricity} of the output, as a core objective. The outputs should evoke novel, interesting metaphors regardless of the domains involved. Second, we want the generated metaphoric outputs to evoke the \textbf{source} domains (eg. ``She \textbf{destroyed} his argument'' evokes the source domain of \source{war}).
        
        We recruited three domain experts in metaphoricity. They were instructed to rate each instance on a scale from 1 (not at all) to 4 (very) for metaphoricity and for whether it evokes the source domain. If the sentence was completely unintelligible, they were instructed to mark it as 0 for both categories. For metaphoricity, annotators were given brief definitions of metaphoricity which they incorporated into their expert knowledge to best rate metaphors. For source domain evocation, they were additionally provided with links to the respective FrameNet frames.
        
        We evaluate three different models for the gold metaphors: the best performing previous model, \textsc{mermaid}, as well as the \textbf{lexical} and \textbf{CM-BART} models. For all models 
        we evaluate generation using the mappings for our gold test set. For the unknown metaphors without gold sentences, we only evaluate our two controlled models, as the generic baselines give the same output regardless of the intended source. This yields a total of 450 sentences (150 gold, 300 without) that are evaluated for metaphoricity and source domain.

        All three experts annotated a random set of 100 training sentences, in order to determine the feasibility and agreement for this task. Agreement rates were .50 for metaphoricity and .37 for source domain (Krippendorff's $\alpha$).\footnote{Full annotation analysis can be found in Appendix \ref{app:anns}.}  

    \subsubsection{Gold Test Mappings}
    
           \begin{table}[t]
        \centering
        \small
        \begin{tabular}{|c|c|c||c|c|c|c|c|c|c|c|c|c|c|}
        \hline
        & \multicolumn{2}{c||}{\textbf{Gold}} & \multicolumn{2}{c|}{\textbf{Rare}} & \multicolumn{2}{c|}{\textbf{Unseen}} \\
        \hline
        \textbf{Model} & \textbf{Met} & \textbf{Src} & \textbf{Met} & \textbf{Src} &  \textbf{Met} & \textbf{Src}  \\
        \hline
        MetMask & 2.27 & 1.60 & - & - & - & - \\
        \textsc{mermaid} & 2.56 & 2.12 & - & - & - & - \\
        \hline
        CM-Lex  & 2.34 & 2.43 & 2.28 & 2.10 & 1.58 & 1.14 \\
       
        CM-BART & \textbf{2.72} & \textbf{2.87} &\textbf{2.41} & \textbf{2.70} & \textbf{2.41} & \textbf{2.01}  \\

        \hline
        \end{tabular}
        \caption{Human evaluations for metaphoricity (\textbf{Met}) and source domain evocation (\textbf{Src}).}
        \label{tab:human-eval}
        \vspace{-4ex}
    \end{table}

    \begin{table*}[tb]
    \small
\begin{center}
\begin{tabular} {|c|c|c|c|c|c|}
    \hline
     & \textbf{Input}/\target{target}/\source{source} & \textbf{Model} & \textbf{Output} & Met & Src \\
    \hline
    \hline


\multirow{5}{*}{1} & \multirow{5}{*}{\makecell{He \textbf{resisted} the panic of vertigo \\ \target{self\_control} \textsc{is} \\ \source{quarreling}}} & Gold  & He \textbf{fought} the panic of vertigo & &  \\
    \cline{3-6}
    & & MetMask & He got the panic of vertigo & 3 & 1  \\
    & & \textsc{mermaid} & He felt the panic of vertigo & 1 & 2  \\    
    & & CM-Lex & He confrontations the panic of vertigo & 0 & 0 \\
    & & CM-BART & He disputed the panic of vertigo & 3 &  4 \\
    \hline

\multirow{5}{*}{2} & \multirow{5}{*}{\makecell{A dim aurora \textbf{rises} in my east \\ \target{change\_position\_on\_a\_scale} \\  \textsc{is} \source{residence}}} & Gold  & A dim aurora \textbf{lives} in my east &  &   \\
    \cline{3-6}
    & & MetMask & A dim aurora kicked in my east & 3  & 1   \\
    & & \textsc{mermaid} & A dim aurora hangs in my east & 4 & 2  \\    
    & & CM-Lex & A dim aurora stands in my east & 3 & 3 \\
    & & CM-BART & A dim aurora lives in my east & 3 & 4  \\
    \hline

\multirow{5}{*}{3} & \multirow{5}{*}{\makecell{People were \textbf{running} out of the theater \\ \target{self\_motion} \textsc{is} \\ \source{fluidic\_motion}}} & Gold  & People were \textbf{streaming} out of the theater &  &   \\
    \cline{3-6}
    & & MetMask & People were clogged out of the theater & 4 & 1  \\
    & & \textsc{mermaid}  & People were running out of the theater & 1 & 4 \\    
    & & CM-Lex & People were boiling out of the theater & 4 & 4 \\
    & & CM-BART & People were spilled out of the theater & 4 & 3 \\
    \hline


    \end{tabular}
\end{center}
\vspace{-.5em}
\caption{\label{tab:result-examples}Example outputs of each system along with the mean of their human evaluations.}
\end{table*}

   \begin{table*}[tb]
    \small
\begin{center}
\begin{tabular} {|c|c|c|c|c|c|}
    \hline
     & \target{target}/\source{source} & \textbf{Model} & \textbf{Output} & Met & Src \\
    \hline
     \multirow{5}{*}{1} & \target{operate\_vehicle} \textsc{is}   & Input & The car \textbf{drove} up alongside him  & &  \\
         \cline{2-6}
    & \multirow{2}{*}{\makecell{Rare: \source{self\_motion}}}
    & CM-Lex & The car drove up alongside him & 1 & 1 \\
    & & CM-BART & The car ran up alongside him & 4 & 4\\
    \cline{2-6}
    & \multirow{2}{*}{\makecell{Unseen: \source{death}}}
    & CM-Lex & The car fell up alongside him & 4 & 4  \\
    & & CM-BART & The car died up alongside him & 4 & 2\\
    \hline
    



\multirow{5}{*}{2} & \target{distributed\_position} \textsc{is} & Input & The meat was \textbf{covered} in a fatty gravy & & \\
    \cline{2-6}
& \multirow{2}{*}{\makecell{Rare: \source{giving}}}
    & CM-Lex & The meat was raised in a fatty gravy & 4 & 1  \\
    &  & CM-BART & The meat was given in a fatty gravy & 2 & 4\\
    \cline{2-6}
& \multirow{2}{*}{\makecell{Unseen: \source{surrendering\_possession}}}
    & CM-Lex & The meat was cut in a fatty gravy & 1 & 1  \\
    & & CM-BART & The meat was yielded in a fatty gravy & 3 & 4\\
    \hline
  
\multirow{5}{*}{3} & \target{dispersal} \textsc{is} & Input & At last the darkness began to \textbf{dissolve} & & \\
    \cline{2-6}
& \multirow{2}{*}{\makecell{Rare: \source{attempt}}}
    & CM-Lex & At last the darkness began to gorn & 0 & 0  \\
    & & CM-BART & At last the darkness began to try & 4 & 4 \\
    \cline{2-6}
& \multirow{2}{*}{\makecell{Unseen: \source{warning}}}
    & CM-Lex & At last the darkness began to Giffen & 0 & 0  \\
    & & CM-BART & At last the darkness began to bite & 4 & 1\\
    \hline

    \end{tabular}
\end{center}
\vspace{-.5em}
\caption{\label{tab:result-unseen}Examples of system outputs on rare and unknown metaphoric mappings.}
\vspace{-1.5em}
\end{table*}

    Results for human evaluations of gold, rare, and unseen metaphoric mappings are shown in Table \ref{tab:human-eval}. With regard to the gold mappings, the CM-BART model performs best in metaphoricity and source domain evocation. CM-Lex has middling performance for metaphoricity, but does well at generating correct source domains. The \textsc{mermaid} system performs well in terms of metaphor generation, but fails to capture the intended source domain.

    Examples of each model's generation are shown in Table \ref{tab:result-examples}. In 1, we see that CM-Lex generates noise, making the results unintelligible. CM-BART  is more robust, generating fluent expressions, and shows evidence of conceptual mapping control,  generating a metaphoric expression matching the source domain. In 2, the MetMask and \textsc{mermaid} models generate reasonable metaphors, which do not evoke the intended domain. CM-Lex is better, generating ``stand'' which can reflect \source{residence}, while the CM-BART performs best, generating the gold metaphoric expression. 

    In 3, we see that the unconstrained models generate effective expressions: "clog" is an evocative metaphor, and "running", while literal, can match the intended domain via the idea of running water. However, our controlled methods both generate novel metaphors that directly evoke the source domain, showing the effectiveness of incorporating conceptual information in generation.

    Overall, we see that the unconstrained models often generate good metaphors, but lack consistency with the input, as they are naive with regard to the conceptual backing of these metaphoric expressions. CM-Lex is effective to some degree, even without metaphoric training data, and CM-BART performs best, generating novel metaphors that frequently match the intended metaphoric expression. 
    
    \subsubsection{Unknown Metaphor Mappings}
    CM-BART outperforms CM-Lex for metaphoricity and source domain evocation for rare and unseen source domains. Examples of the two proposed models' generated for rare and unseen metaphoric mappings are shown in Table \ref{tab:result-unseen}.

    Example 1 shows the ideal case. When given a source domain from a "rare" mapping, the resulting metaphor is fairly reasonable. CM-BART generates a metaphor consistent with the original semantics; CM-Lex generates the literal utterance. When presented with an unseen mapping in which operating a vehicle is framed as death, we get diverse expressions, both adding meaning to the original utterance. CM-Lex uses the verb "fell" (albeit incorrectly conjugated), which can be used to abstractly evoke the death domain, while CM-BART directly uses the verb "die". The original expression can be ambiguous as to whether the car stopped: the evoked metaphor enforces the stoppage of the car, and also provides color to the expression.
    
    Example 3 highlights a key issue: when the source and target domains are too incongruent, the generated expressions can be inconsistent. CM-Lex here again generates noise. However, CM-BART generates normal, expressive metaphors, which are nonetheless incompatible with the original literal input, which denotes the lessening of darkness. Rather, CM-BART generates a metaphor expressing perhaps growing darkness with the verb \textbf{try} and a dangerous darkness with the verb \textbf{bite}.
    
    This is a critical point with regard to conceptual mappings. Not all pairs are available: they require semantic consistency, and while generating from any two pairs may yield insightful, interesting, and perhaps inspiring new metaphoric expressions, generating metaphoric paraphrases requires additional knowledge of which source/target pairings are compatible. This generally supports notion of invariance and structure mapping, in which there is inherent structure within domains that needs to be consistent in order to  evoke metaphoric mappings between them \cite{gentner1983,lakoff1993}.
    
    It must be noted that the systems proposed here have a distinct advantage in this task: we add FrameNet frames, which, while neither perfect nor designed to capture metaphoricity, provide a strong signal for which domains to generate in. This highlights a possible benefit to the interaction between deep, pre-trained models such as BART and available lexical resources: by combining these, we are able to leverage the strength of each to build a powerful metaphor generation system.

\section{Related Work}
 
    We broadly cover two areas of related work: previous computational approaches to CMT, and previous approaches to metaphor generation.
    
\paragraph{Computational Approaches to CMT.}
       
    There are a variety of approaches to identifying conceptual metaphors themselves. The CorMet system \cite{mason2004} was built to extract conceptual metaphors based on selectional preferences of verbs. \newcite{shaikh2014} builds "conceptual spaces" for source domains, using rule-based extraction of relations between lexical items. These conceptual spaces are then used to find new conceptual metaphors. This process is extended to build a repository of linguistic and conceptual metaphors \cite{shaikh2014-2}. \newcite{mohler2014} focus on identifying appropriate source domains for metaphoric expressions, using vector-based approaches for metaphor interpretation. 
      
    The idea of using frames to represent metaphoric domains has been explored in the MetaNet project \cite{dodge2015}. We however, restrict our work to FrameNet due to the coverage and availability of reliable automatic parsing.

\paragraph{Metaphor Generation.}
Early work in metaphor generation was based in heuristics, learning to generate relatively simple "A is like B" representations \cite{abe2006,terai2010}. In a similar vein, \newcite{veale2016} uses template-like structures to generate creative and metaphoric tweets.

Other works focus on identifying metaphoric mappings using WordNet clustering and selectional preferences \cite{mason2004,gandy2013}, syntactic relations to build proposition databases \cite{ovchinnikova2014}, and embedding based approaches to identify poetic relationships \cite{gagliano2016}. However, the goal of these works is to generate mappings, rather than linguistic expressions that evoke them.

Amongst deep learning approaches \newcite{yu2019} identify literal and metaphoric words in corpora based on selectional restrictions, and using these to train sequence-to-sequence models for metaphor generation, albeit without reference to any input expression. \newcite{stowe2020} generates metaphors using masked language modeling, masking metaphoric tokens in training in order to encourage metaphoric generation. Other approaches use novel methods for collecting literal/metaphor pairs, training sequence-to-sequence models for simile generation and metaphoric paraphrasing \cite{chakrabarty2020,chakrabarty2021}. These approaches effectively generate figurative language, but the models have no knowledge of the underlying metaphors, and thus simply generate ungrounded expressions. This leads to outputs which are possibly metaphoric, but contain no connection to the input, eschewing the critical connections that make novel metaphors powerful. We instead propose methods for generating metaphoric paraphrases grounded in CMT. 

\section{Conclusions and Future Work}
    In summary, we have shown two methods for incorporating knowledge of conceptual metaphor theory in metaphor generation. We trained FrameNet frame embeddings to represent conceptual domains, and applied shifts between them to generate metaphors in an unsupervised fashion. Leveraging FrameNet further, we build a dataset of semi-supervised pairs that evoke conceptual metaphors, which can be used along with BART for controlled metaphor generation. This model achieves state-of-the-art performance in metaphor generation by both automatic and human evaluations.
    
    Future work can expand these models to go beyond verbs, incorporating nominal and other types of metaphors. The next necessary step is to go beyond lexicalized metaphors: good, consistent conceptual metaphors often span long stretches of text, and we need to design models that can learn and generate metaphors over larger texts.
    
\section*{Ethical Considerations}
Although we use language models trained on data collected from the Web, which have been shown to have issues with bias and abusive language  \cite{sheng-etal-2019-woman, wallace-etal-2019-universal}, the inductive bias of our models should limit inadvertent negative impacts. Unlike model variants such as GPT, BART is a conditional language model, which provides more control of the generated output. It should also be noted that  our CM-BART model is fine-tuned on the poetry corpus which is devoid of harmful and toxic text especially targeted at marginalized communities

Advances in generative AI inherently come with concerns about models' ability to deceive, persuade, and misinform. Metaphorical language has been shown to express and elicit stronger emotion than literal language \cite{citron2014metaphorical,mohammad2016} and to provoke emotional responses in the context of political discourse covered by mainstream newspapers \cite{figar2014emotional}. We understand there may be concerns about building generative models for metaphors aimed at persuasion. Social scientists distinguish persuasion from manipulation based on two aspects: dissimulation and constraint \cite{Nettel2012}.  Dissimulation involves concealing intention, which requires hiding information, whereas constraint involves removing options from the audience and forcing them to accept the conclusion. Our work on metaphor generation  does not aim to hide information about a topic or present it as the only choice, but aims to provide the same sentence using more expressive language.

\bibliography{acl2021}
\bibliographystyle{acl_natbib}

\appendix

\pagebreak
\section{Appendix A}
    
    \begin{table*}[ht]
        \centering
        \begin{tabular}{|c|c|c|c|c|c|c|c|c|c|}
        \hline
         & \multicolumn{3}{c|}{$lex$ sim} & \multicolumn{3}{c|}{$str$ sim} & \multicolumn{3}{c|}{mean} \\
         \cline{2-10}
        Dimensions
         & 50 & 100 & 300 & 50 & 100 & 300 & 50 & 100 & 300 \\
        \hline
        \hline
        \texttt{word2vec} &.203 & \textbf{.208} & .205 & \textbf{.111} & .076 & .104 & \textbf{.157} & .144 & .154\\
        \texttt{fasttext} & .113 & .120 & .117 & .042 & .103 & .095 & .077  &.111 & .106 \\
        \texttt{glove} & .179 & .191 & .212 & -.106 & -.136 & -.108 & .037 & .028 & .052 \\
        \hline
      
        \end{tabular}
        \caption{}
        \label{tab:frame_embeddings}
    \end{table*}

\label{app:metrics}
Results for each frame embedding method using the distance metrics defined in Section \ref{sec:framemethod} are shown in Table \ref{tab:frame_embeddings}.
     \begin{figure}[t]
    \includegraphics[width=.47\textwidth]{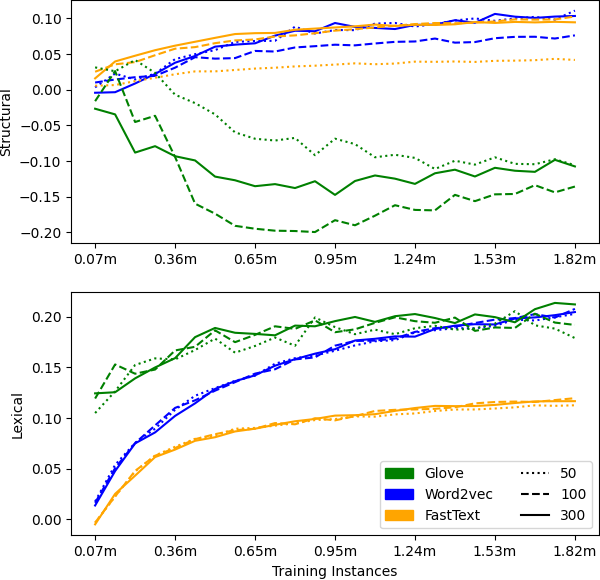}
    \centering
    \vspace{-1ex}
    \caption{\label{fig:frame_embeddings} Frame embedding evaluation metrics as data is added.}
    \end{figure}
    
    \begin{table}[]
        \centering
        \small
        \begin{tabular}{c|c|c|c}
        \hline
       Paper & n & Method & Agreement \\
       \hline
              \newcite{dodinh2018} & 15,180 & MTurk & .16-.38 $\alpha$ \\
       \hline
       \newcite{yu2019} & 80 & MTurk & - \\
       \newcite{chakrabarty2020} & 900 & MTurk & .36-.49 $\alpha$ \\
       \newcite{stowe2020} & 513 & MTurk & - \\
       \newcite{chakrabarty2021} & 900 & MTurk  & - \\
       \hline
       This work & 450 & Experts & .505 $\alpha$ \\
        \end{tabular}
       \caption{Comparison of agreement rates for various metaphor evaluation tasks. Note that \newcite{dodinh2018} developed a real-valued scoring layer over an existing corpus rather than evaluating generated outputs. ``-'' indicates agreement is not reported.}
        \label{tab:agreement}
    \end{table}

  Figure \ref{fig:frame_embeddings} tracks these evaluation metrics as more data is added to each algorithm. The lexical evaluation relatively stable, peaking in most cases between .1 and .2. The word2vec embeddings maintain their upward progression even at maximal data: theoretically additional data could improve these embeddings further.  The structural evaluation shows something very different: while word2vec and FastText embeddings improve as data is added, showing some effects of model size, the Glove embeddings trend sharply negative at first before proceeding beginning to improve.

\section{Appendix B}
\label{app:anns}

 Agreement rates were measured using Krippendorff's $\alpha$. For metaphoricity, the mean score was .505, indicating moderate agreement. However, given the difficulty of this task, we believe this to be relatively stronger: see Table \ref{tab:agreement} for comparison to other work evaluating metaphor generation.
 
 For source domain annotation, annotators varied in the degree to which source domains were evoked. Initial agreement was relatively poor (.249): we performed a post-processing step, normalizing their results to a consistent mean. This yields an agreement score of .387: which we deemed competitive for the difficulty of the task. As we have no direct comparison for evaluation, further work is required to refine this type of evaluation process.
 
 \section{Appendix C}
 \label{app:params}
 For retrieving commonsense symbolism of the sentences, we use the pre-trained COMET model \footnote{\url{https://github.com/atcbosselut/comet-commonsense}} and retrieve top 5 candidates for each input.
\begin{enumerate}[noitemsep]
    \item{\textbf{No of Parameters:}} We use the BART large checkpoint (400M parameters) and use the FAIRSEQ implementation \cite{ott2019}
     \footnote{\url{https://github.com/pytorch/fairseq/tree/master/examples/bart}}.
    \item{\textbf{No of Epochs:}} We fine-tune pre-trained BART for 25 epochs for CM-BART model and save the best model based on validation perplexity. 
    \item{\textbf{Training Time:}} Our training time is 60 minutes for CM-BART.
    \item{\textbf{Hardware Configuration:}} We use 4 RTX 2080 GPUs.
    \item{\textbf{Training Hyper parameters:}} We use the same parameters as the FAIRSEQ github repository where BART was fine-tuned for the CNN-DM summarization task with the exception of the size of each mini-batch, in terms of the number of tokens, for which we used 1024. \footnote{\url{https://github.com/pytorch/fairseq/blob/master/examples/roberta/README.glue.md}}
    \item{\textbf{Decoding Strategy \& Hyper Parameters:}} For decoding we generate metaphors from our models using a top-k random sampling scheme \cite{fan2018}. At each timestep, the model generates the probability of each word in the vocabulary being the likely next word. We randomly sample from the k = 5 most likely candidates from this distribution.

\end{enumerate}

\end{document}